\documentclass[runningheads]{llncs}

\usepackage{amsmath}
\usepackage[american]{babel}
\usepackage{microtype}
\usepackage{graphicx}
\usepackage{algorithm}
\usepackage{algorithmic}
\usepackage{siunitx}

\usepackage[caption=false]{subfig}

\usepackage{float}
\usepackage{pdfpages}
\usepackage{multirow}
\usepackage{mathtools}
\usepackage{amsmath}
\usepackage{amssymb}
\usepackage{cite}
\usepackage{siunitx}

\DeclareMathOperator*{\argmax}{arg\,max}
\DeclareMathOperator*{\argmin}{arg\,min}

\begin{document}

\title{Unifying Optimization Methods for Color Filter Design \thanks{Supported by EPSRC under Grant EP/S028730 and Apple Inc.}}
\author{Graham Finlayson\inst{} \and
Yuteng Zhu\inst{*}}
\authorrunning{G. Finlayson and Y. Zhu}
\institute{ {\large (co-first authors)} \\ 
University of East Anglia, Norwich NR4 7TJ, UK \\
\email{\{yuteng.zhu\}@uea.ac.uk}}
\maketitle             

\begin{abstract}

Through optimization we can solve for a filter that when the camera views the world through this filter, it is more colorimetric.  Previous work solved for the filter that best satisfied the Luther condition: the camera spectral sensitivities after filtering were approximately a linear transform from the CIE XYZ color matching functions. 
A more recent method optimized for the filter that maximized the Vora-Value (a measure which relates to the closeness of the vector spaces spanned by the camera sensors and human vision sensors). 
The optimized Luther- and Vora-filters are different from one another. 

In this paper we begin by observing that the function defining the Vora-Value is  equivalent to the Luther-condition optimization if we use the orthonormal basis of the XYZ color matching functions, i.e.\ we linearly transform the XYZ sensitivities to a set of orthonormal basis. In this formulation, the Luther-optimization algorithm is shown to almost optimize the Vora-Value.
Moreover, experiments demonstrate that the modified orthonormal Luther-method finds  the same color filter compared to the Vora-Value filter optimization. Significantly, our modified algorithm is simpler in formulation and also converges faster than the direct Vora-Value method.

\end{abstract}

\section{Introduction}\label{sec_intro}

\begin{figure}[ht]
  \centering
  \includegraphics[width=0.75\textwidth]{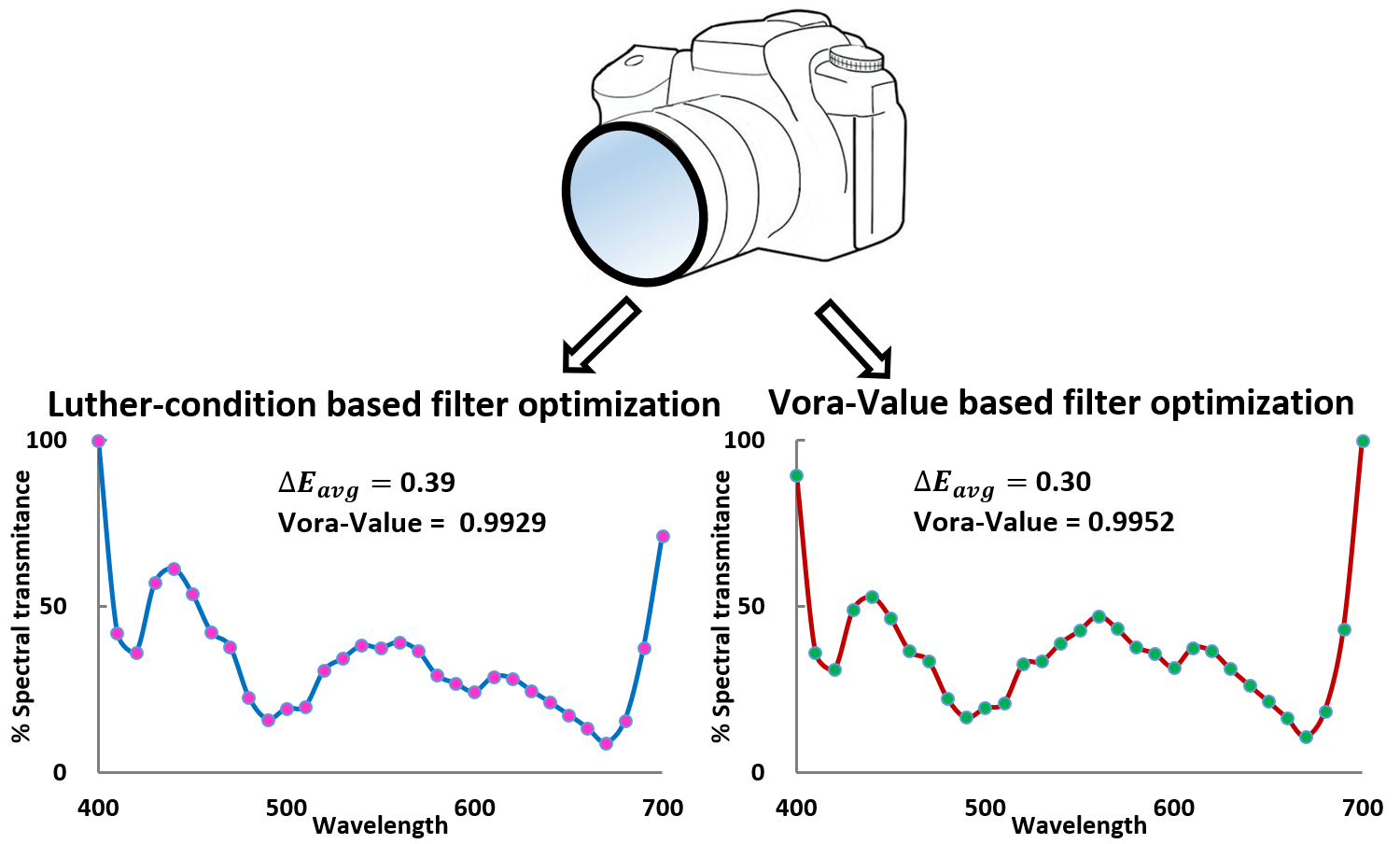} 
  \caption{The upper panel illustrates the general idea of placing a filter in front of a digital camera. The bottom panel shows the optimized filters solved from two filter optimization methods for a Canon 5D Mark II DSLR camera. These two optimized filters have different spectral transmittance and consequently perform differently in terms of averaged CIELAB color error $\Delta E_{ab}^*$ and Vora-Value score.}
  \label{fig:demo}
\end{figure}

A digital camera sees the world through Red, Green and Blue sensors. However, for practical considerations (including manufacturability and the need to have low image noise~\cite{vazquez2014perceptual}), the RGB sensors are not linearly related to the human vision sensitivity functions~\cite{schanda2007colorimetry}.
A camera is said to be colorimetric if it satisfies the so-called Luther condition: its spectral sensitivities are a linear combination of the CIE XYZ color matching functions (CMFs)~\cite{luther1927gebiet}. 
While the Luther condition is never met exactly,  the {\bf closer} the spectral responses of a camera are to being linearly related to the CMFs, the better we can  correct the camera colors to XYZs or to a display RGB space such as sRGB~\cite{berns2019billmeyer}.

The devil is in the detail, specifically in what we mean by `closer'. We need to quantitatively measure `closeness' in some sense.
Neugebauer proposed the `Q-factor'  sensors~\cite{neugebauer1956quality} but this has the weakness that it does not incorporate linear transforms into the measure. This is a serious omission as we always correct the measured camera RGBs - apply a linear transform - to make an image suitable for display. Later, the \textit{Vora-Value} was proposed by Vora and Trussell ~\cite{vora1993measure}. The Vora-Value is a carefully crafted formula that both admits  the idea of linear transform and also is designed to quantify sensor performance given all theoretically possible light stimuli. The Vora-Value is the measure of `goodness' we adopt in this paper.
Related to this paper, prior work has shown how the Vora-Value~\cite{vora1997design} can, at least in principle, be used as a criterion with respect to which we might design the spectral sensitivities in camera. However, this is a theoretical result: the optimized filters cannot be easily manufactured.

Rather than trying to make new spectral sensitivities, in this paper, we propose to make an off-the-shelf camera more colorimetric by placing a filter in front of the camera (see Fig. 1). The filter is designed so that the filtered RGBs are approximately linearly related to the reference XYZs.  Previous work by Finlayson \emph{et al.}~\cite{CIC2018} optimized for the filter to meet the  Luther condition~\cite{luther1927gebiet}. Specifically, they solve for the spectral transmittance of the color filter that when  multiplied by the camera sensitivity functions is almost a linear transform from the CIE color matching functions.  The best filter is found via a simple and intuitive alternating least-squares algorithm where the filter and the mapping transformation solutions are solved in turn until convergence. The resulting Luther-filter for a Canon 5D Mark II DLSR camera is shown at the bottom left of Fig. 1.
Arguably, a weakness of the method is that it is tied to a single fixed set of color matching functions. Indeed, if we solve for the best filter for  the XYZs CMFs and the sRGB matching functions (for Rec 709 primaries~\cite{itu2002parameter}) then we obtain different filters.

 Recent work solved for - using a gradient ascent algorithm - the filter that maximized the Vora-Value criterion~\cite{CIC2020}. By formulation, the Vora-Value is based on the `vector space' idea and for given camera and XYZ sensitivities it calculates the same value irrespective of any linear transform of these two sensor sets.
The Vora-Value designed filter is shown at the bottom right of Fig. 1. Interested readers are referred to~\cite{CIC2020} for more detail about the optimization.  

Depending on the optimization criterion at hand, we recover different filters. Unsurprisingly, `scored' for the closeness to XYZs in the Luther-optimization, the Luther-filter achieves a better fit to CMFs than the Vora-Value filter. Conversely, the Vora-Value optimized filter achieves a higher Vora-Value score compared to the Luther-filter. This said, the algorithm for the Luther-condition approach is much simpler than the Vora-Value approach. The latter maximizes the Vora-Value by a Gradient Ascent approach. Not only is the form of the gradient function 
complex but the gradient search is slow to converge. In contrast, the Luther-condition method is simple and can be solved by the ALS algorithm with fast convergence speed.
An additional advantage of the ALS approach is that it is straightforward to extend to incorporate measured lights and reflectances~\cite{2019CCIW,2020EI}.

In this paper, we modify the simpler Luther-optimization so it can be used to optimize the Vora-Value. We revisit the formulation of the Vora-Value and show that it is more simply expressed when we map the XYZ CMFs to the corresponding orthonormal basis. The ALS method can be used to find the filter so that the filtered camera sensitivities that are linearly related to a linear combination of XYZ CMFs (which is orthonormal). By solving the modified Luther condition, we can show we must also be optimizing the Vora-Value. 

Experiments demonstrate two important results. First, for all cameras tested, we find the same filter using the modified Luther optimization and the original gradient-ascent Vora-Value optimization. Second, the convergence speed is significantly faster using our new modified Luther-condition method.

The rest of the paper is organized as follows. Section~\ref{sec_background} reviews the definition of the Luther condition and Vora-Value, and also introduces the alternating least-squares filter design
algorithm. In Section~\ref{sec_design}, we present the modified Luther-condition optimization and prove its equivalence to the Vora-Value optimization. Later we show how the ALS algorithm can be used to find the filter that maximizes the Vora-Value.  Experimental results are presented in Section~\ref{sec_exp}. The paper concludes in Section~\ref{sec_conclusion}.

\section{Background} \label{sec_background}
\subsection{The Luther Condition}

The Luther condition states that the spectral sensitivity curves of the camera sensors are a linear combination of the CIE XYZ color matching functions (CMFs). 
Let $\mathbf{Q} = \mathbf{[r,g,b]}$ and  $\mathbf{X} = \mathbf{[x,y,z]}$ denote respectively the spectral sensitivities of the camera and the CMFs of the human visual sensors. The columns of matrices $\mathbf{Q}$ and  $\mathbf{X}$ represent the spectral sensitivity for each sensor channel and the rows represent the sensor responses at a sampled wavelength. Both matrices are in the size of $n \times 3$, where $n$ is the number of sampling wavelengths across the visible spectrum (typically, $n=31$ when the visible spectrum running from 400\,nm to 700\,nm is sampled every 10\,nm).

Mathematically, the Luther condition is written as
\begin{equation}
    \mathbf{X = QM} 
    \label{eq:Luther}
\end{equation}
where $\mathbf{M}$ is a $3 \times 3$  matrix.

Equivalently we write:
\begin{equation}
    \mathbf{XT_1 = QT_2M'} 
    \label{eq:Luther1}
\end{equation}

\noindent 
where $\mathbf{T_1}$, $\mathbf{T_2}$ and $\mathbf{M'}$ are full rank $3\times 3$ matrices.
Clearly, given Eq.~(\ref{eq:Luther}), then $\mathbf{M'=[T_2]^{-1}MT_1}$. That is, the Luther condition does not depend on the particular basis used (we can map camera sensitivities to XYZs, cone functions or any linear combination thereof). 

If a camera satisfies the Luther condition, for any two color signals, $\mathbf{f}$ and $\mathbf{g}$ that produce the same values by the camera sensors, they should also make the same XYZ stimulus values. That is,
\begin{equation}
  \mathbf{ Q^T{f} = Q^T{g} \; \xRightarrow{M} \; 
   X^T{f}= X^T{g}}
\end{equation}
and thus are indistinguishable to the human observer. Readers are referred to ~\cite{HORN1984} for more detail.

\subsection{The Vora-Value}
The Vora-Value is often used to measure  how similarly a camera samples the spectral world compared to the human visual system. The Vora-Value is a number between 0 and 1 where 1 means the camera is fully colorimetric such that RGBs are precisely a linear transform from XYZ tristimulus values.

Given a camera sensor set $\mathbf{Q}$ and the human visual sensors $\mathbf{X}$, the Vora-Value is defined~\cite{vora1993measure} as
\begin{equation}
\begin{split}
     \mathbf{ \nu (Q,X)} 
     =\frac{1}{3}trace(\mathrm{P}\{\mathbf{Q}\}\mathrm{P}\{\mathbf{X}\})
\end{split}
\label{eq:vv_def}
\end{equation}
where $\mathrm{P}\{\mathbf{Q}\}$ and $\mathrm{P}\{\mathbf{X}\}$ denote the projection matrices of the camera spectral responses and the human visual sensitivities, respectively and $trace()$ is the trace of a square matrix that sums up the elements along the diagonal of a matrix. The projector of a matrix - such as $\mathbf{Q}$ - is equal to 
\begin{equation}
    \mathrm{P}\{\mathbf{Q}\}=\mathbf{Q[Q^T Q]^{-1}Q^T}
\end{equation}
where the superscripts $^T$ and $^{-1}$ denote respectively the matrix transpose and inverse.

\subsection{Orthonormal Basis}

Let $\mathbf{U}$ and $\mathbf{V}$ respectively denote linear combinations of $\mathbf{X}$ and $\mathbf{Q}$ that are orthogonal, i.e.\ $\mathbf{U}=\mathbf{QT_1}$ and $\mathbf{V}=\mathbf{XT_2}$ where $\mathbf{U}^T \mathbf{U}=\mathbf{I}_{3}=\mathbf{V}^T \mathbf{V}$ ($\mathbf{I}_{3}$ is the ${3\times 3}$ identity matrix). 

By simple substitution into the matrix projector, we obtain
\begin{equation}
    \mathrm{P}\{\mathbf{Q}\}= \mathrm{P}\{\mathbf{U}\}= \mathbf{UU^T}
\end{equation}
and 
\begin{equation}
\mathrm{P}\{\mathbf{X}\} = \mathrm{P}\{\mathbf{V}\} =\mathbf{VV^T}.
\label{eq:eq7}
\end{equation}
By using Eq.~(\ref{eq:eq7}) for the Vora-Value definition, we have $\mathbf{ \nu (Q,X)} = \mathbf{ \nu (Q,V)}$.
We will make use of the orthonormal bases in the next section.

\subsection{Least-squares Correction and Filtering}
First, let us begin with the simple least-squares (LS) regression case. Given two matrices, e.g.\ $\mathbf{X} $ and  $\mathbf{Q}$ (in the size of $m \times n$ with $m \geq n$), the best predictor of $\mathbf{X}$ from $\mathbf{Q}$ can be found in the closed-form  by  
\begin{equation}
    \mathbf{M = [Q^T Q]^{-1} Q^T X}
    \label{eq:LS}
\end{equation}
where $\mathbf{X\approx QM}$ gives the least sum-of-squared errors.

Physically, the effect of placing a transmissive filter $\mathbf{f}$ (a $31\times 1$ vector) in front of a camera is written as $diag(\mathbf{f})\mathbf{Q}$ where $diag()$ is a function that turns a vector into a diagonal matrix. To ease notation by $\mathbf{F}=diag(\mathbf{f})$, so the filter multiplied by camera is written as $\mathbf{FQ}$. For a given $\mathbf{Q}$, we find the best $\mathbf F$ row-by-row and in closed form:

\begin{equation}
\mathbf F_{ii}=\frac{\mathbf Q_i\cdot \mathbf X_i}{\mathbf Q_i \cdot \mathbf Q_i}
\label{eq:LS1}
\end{equation}
where the subscript $i$ denotes the $i$th row of a matrix and `$\cdot$' is the vector dot-product. 

In the Luther-condition optimization - recapitulated below - we need to simultaneously solve for the best $\mathbf M$ and $\mathbf F$ that together map the camera sensitivities as close as possible to $\mathbf X$ as
\begin{equation}
	\argmin \limits_{ \mathbf{F,M}}\parallel{ \mathbf{FQM -  X}}\parallel^2_2.
\label{eq:LutherOptOrig}
\end{equation}

\section{The Modified Luther-condition Optimization}\label{sec_design}

We propose a simple modification to the Luther-condition optimization. We simply substitute the target color matching sensitivities $\mathbf X$ with a special linear transform $\mathbf{V=XT}$ where $\mathbf V$ is chosen to be an orthogonal matrix. Now we optimize:

\begin{equation}
	\argmin \limits_{ \mathbf{F,\; M}}\parallel{ \mathbf{FQM -  V}}\parallel^2_2.
\label{eq:LutherOpt}
\end{equation}

This modified Luther-condition optimization aims for the best filter matrix $\mathbf F$ and the $3 \times3 $ linear transform $\mathbf M$ that return the least errors between the two spectral sensitivities sets.  

Given the filter matrix $\mathbf{F}$, in the least-squares sense, the best $\mathbf{M}$ is obtained 
\begin{equation}
    \mathbf{M = ((FQ)^T FQ)^{-1} (FQ)^T V}
    \label{eq:M4FQ}
\end{equation}
Here we see that the best linear mapping $\mathbf M$ in the Luther-condition optimization of Eq.~(\ref{eq:LutherOpt}) is essentially a function of $\mathbf{F}$.

By multiplying with $\mathbf{FQ}$, we have 
\begin{equation}
   \mathbf{ FQM = FQ((FQ)^T FQ)^{-1} (FQ)^T V }= \mathrm{P}\{\mathbf{FQ}\}V
    \label{eq:FQM}
\end{equation}
Substituting into Eq.~(\ref{eq:LutherOpt}), the optimization can be rewritten as
\begin{equation}
	\argmin \limits_{\mathbf F}\parallel{(\mathrm{P}\{\mathbf{FQ}\}-\mathbf{I)  V}}\parallel^2_2
\label{eq:LutherF}
\end{equation}
where $\mathbf I$ is the identity matrix. 

\begin{theorem}
By minimizing $\argmin\limits_{\mathbf F}\parallel{(\mathrm{P}\{\mathbf{FQ}\}-\mathbf{I)  V}}\parallel^2_2$, we maximize $v(\mathbf{FQ},\mathbf{X})$.
\end{theorem}

\noindent
{\it Proof:} 

\noindent
We will use the following rules:
\begin{enumerate}
\item $trace(\mathbf{A^TA})= ||\mathbf{A}||^2_2$ (remember, trace is the sum of the diagonal of a matrix)
\item $trace(\mathbf{ABC})=trace(\mathbf{BCA})=trace(\mathbf{CBA})$, the acyclic of matrix trace
\item $trace(\mathbf{A+B})=trace(\mathbf{A}) +trace(\mathbf{B})$
\item $trace(\mathrm{P}\{\mathbf{A}\})=rank(\mathbf A)$, the trace of a projector is the rank of the matrix
\item $\mathrm{P}\{\mathbf{A}\}\mathrm{P}\{\mathbf{A}\}=\mathrm{P}\{\mathbf{A}\}$, idempotency of the projector
\item $\mathrm{P}\{\mathbf{A}\} = (\mathrm{P}\{\mathbf{A}\})^T$, the projector is symmetric.
\end{enumerate}

Using the property of rule 1, we can rewrite the formula in Eq.~(\ref{eq:LutherF})  as 
\begin{equation}
    \parallel{(\mathrm{P}\{\mathbf{FQ}\}-\mathbf{I)  V}}\parallel^2_2 \; = trace(\mathbf{V}^T(\mathrm{P}\{\mathbf{FQ}\}-\mathbf{I})^T (\mathrm{P}\{\mathbf{FQ}\}-\mathbf{I})\mathbf{V})
\end{equation}

Using rule 2 for moving matrix $\mathbf V^T$, we have
\begin{equation}
    \parallel{(\mathrm{P}\{\mathbf{FQ}\}-\mathbf{I)  V}}\parallel^2_2 \; = trace((\mathrm{P}\{\mathbf{FQ}\}-\mathbf{I})^T (\mathrm{P}\{\mathbf{FQ}\}-\mathbf{I})\mathbf{V}\mathbf{V}^T)
\end{equation}

As known in Eq.~(\ref{eq:eq7}) that $\mathbf{VV^T} = \mathrm{P}\{\mathbf X\}$, we have 
\begin{equation}
    \parallel{(\mathrm{P}\{\mathbf{FQ}\}-\mathbf{I)  V}}\parallel^2_2 \; = trace((\mathrm{P}\{\mathbf{FQ}\}-\mathbf{I}) (\mathrm{P}\{\mathbf{FQ}\}-\mathbf{I})\mathrm{P}\{\mathbf{X}\} )
    \label{proof_step_8}
\end{equation}

Expanding the equation and using rules 5 and 6, we obtain
\begin{equation}
\begin{split}
     \parallel{(\mathrm{P}\{\mathbf{FQ}\}-\mathbf{I)  V}}\parallel^2_2 &= 
    trace((\mathrm{P}\{\mathbf{FQ}\}-2\mathrm{P}\{\mathbf{FQ}\}+I)\mathrm{P}\{\mathbf{X}\}) \\
    &= -trace(\mathrm{P}\{\mathbf{FQ}\}\mathrm{P}\{\mathbf{X}\})+ trace(\mathrm{P}\{\mathbf{X}\})
\end{split}
\end{equation}

Remember the Vora-Value definition in Eq.~(\ref{eq:vv_def}) (and also rule 4), we obtain 
\begin{equation}
    \parallel{(\mathrm{P}\{\mathbf{FQ}\}-\mathbf{I)  V}}\parallel^2_2 \; = -3\,\nu(\mathbf{FQ,X}) + 3
\end{equation}

Now we can prove the equivalence:
\begin{equation}
     \argmin \limits_{\mathbf F}\parallel{(\mathrm{P}\{\mathbf{FQ}\}-\mathbf{I)  V}}\parallel^2_2 \iff \argmax_{\mathbf{F}}\; \mathbf{\nu (FQ, X)}
\end{equation}
as well as
\begin{equation}
     \argmin \limits_{\mathbf {F,\; M}}\parallel{\mathbf{FQM - V}}\parallel^2_2 \iff \argmax_{\mathbf{F}}\; \mathbf{\nu (FQ, X)}
\end{equation}

\subsection{Alternating Least-Squares Algorithm}

To minimize Eq.~(\ref{eq:LutherOpt}), or via Theorem 1 to maximize the Vora-Value, we adopt an alternating least-squares algorithm:

\begin{algorithm}[h]
	\caption{ALS algorithm solving for the filter optimization}
	\label{algo1}
	\begin{algorithmic}[1]
		\STATE{$i=0, F^{0}=diag(\underline{f}^{initial}), Q^{0}=F^{0}Q $}
		\REPEAT
		\STATE{$i = i+1$}
		\STATE{$\min\limits_{M^{i}} \parallel F^{i-1}QM^{i} - \text{\it V}\parallel_{2}^2$}
		\label{codeM}
		\STATE{$\min\limits_{F^{i}} \parallel F^{i}QM^{i} - \text{\it V}\parallel_{2}^2$}
		\label{codeF}
		\STATE{$Q^{i} = F^{i}Q$}
		\UNTIL{$ \nu(Q^{i},V) - \nu(Q^{i-1},V) \, < \, \epsilon $} \label{code:convergence}\\
		\STATE{$ F = F^i $\quad and \quad $M = M^{i} $}\\
		\label{algo1_finalstep}
	\end{algorithmic}
\end{algorithm}

Specifically, starting from an initial filter solution $\mathbf{F^0}$, we first solve for the matrix $\mathbf{M}$ by holding the filter $\mathbf{F}$ fixed (see step~\ref{codeM}) and alternatively using the newly calculated $\mathbf{M}$ to solve for the filter matrix $\mathbf{F}$ (see step~\ref{codeF}) and the process will continue updating both matrices in turn until it converges to a stopping criterion (see step~\ref{code:convergence}). The ALS method is guaranteed to converge (although not necessarily to the global optimum)~\cite{zhang2001rank}.

Steps~\ref{codeM} and~\ref{codeF} - where we find the linear transform and the filter - are solved using simple, closed-form least-squares estimation, see Eqs. (8) and (9).

In~\cite{CIC2020}, the Vora-Value minimization was cast as a gradient ascent optimization. To implement that approach, it was found that the gradient step is complex in formulation and the gradient ascent converged slowly. 
It is important to note that the ALS and gradient ascent algorithm have the same starting optimization statement (that is what we have proved in Theorem 1).  Moreover, both approaches are guaranteed to converge to a fixed point. However, this does not mean that these two minimizations must converge to the same filter solution.

\begin{figure}[!th]
    \centering
    \includegraphics[width=0.75\textwidth]{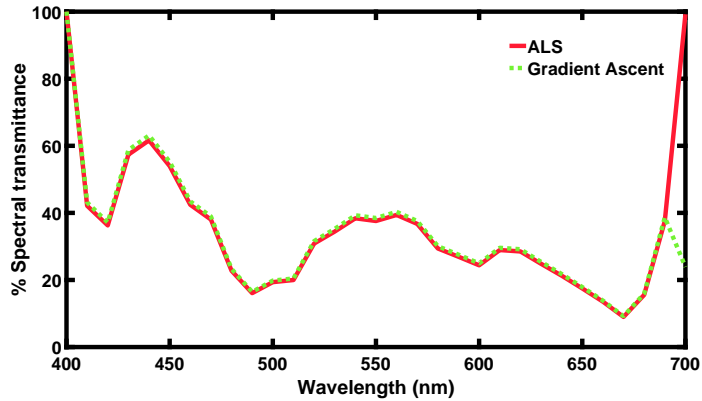}
    \caption{Spectral transmittance of the optimized filters for Canon 5D Mark II camera. The optimized filter in solid red line is solved by ALS algorithm while the filter in dotted green line is solved by the gradient ascent algorithm.}
    \label{fig:filters}
\end{figure}

\section{Experiments and Results}\label{sec_exp}

The optimal filter derived from the Luther-condition based optimization - where the target filter set is $V=XT$, where the columns of $V$ are orthonormal vectors -  for a Canon 5D Mark II camera~\cite{jiang2013space} using Algorithm~\ref{algo1} is shown in Fig.~\ref{fig:filters}, see the solid red line. We also plot the optimal filter of the Vora-Value formulated optimization solved by the gradient ascent algorithm developed in~\cite{CIC2020}, see the dotted green line. From the figure, we can see that these two algorithms give almost the same filter solutions (except at the end of the visible spectrum).

In Table 1, we evaluate the derived filters in terms of Vora-Value. The two filter design methods are termed as \textbf{Luther-ALS} and  \textbf{Vora-GA} by its optimization formulation and the corresponding algorithm. 
We also include the results of the native (without a filter) camera sensor as baseline results (denoted \textbf{Baseline}).
From the table, we can see that \textbf{Luther-ALS} and \textbf{Vora-GA} deliver almost the same performance (the difference is very small and improve significantly from 0.9342 of the native camera sensor set to 0.9952 of the filtered camera sensors. 
As a higher Vora-Value indicates greater similarity of the subspaces spanned by the sensitivities of a camera and human visual system, therefore, generally relates to more accurate color measurement~\cite{vora1993measure}.

\begin{table}[!th]
\centering
\caption{Error statistics for Canon 5D Mark II camera}
\vspace{0.1cm}
\renewcommand{\arraystretch}{1.5}
\setlength{\arrayrulewidth}{0.4mm}
\setlength\tabcolsep{4pt} 
\begin{tabular}{lp{1.5cm}lp{1cm}cccccc}
\hline 
\multirow{2}{*}{Method} & Vora-{} &  & \multicolumn{5}{c}{color   errors $\Delta E_{ab}^{*}$}  \\ \cline{4-8} 
&Value &  & mean  & median  & 95\% & 99\%  & max   \\ \hline
Baseline   &0.9342   &   & 1.416  & 0.836   & 4.262 & 11.786 & 29.392  \\ \hline
Luther-ALS &0.9952  &  &0.298 & 0.147  & 0.983 &	2.472  &	10.127  \\
Vora-GA  &0.9952 &  & 0.300	&0.149  &0.986	&2.472	&10.027   \\\hline 
\end{tabular}
\label{tab1}
\end{table}

Now let us evaluate the derived filters 
with respect to a color measurement experiment.
For a collection of 102 illuminants and 1995 refletance spectra~\cite{barnard2002data}, we calculate the RGBs (for the native camera and the camera sensitivities after filtering) and ground-truth XYZs.  The corresponding CIELAB color difference metric $\Delta E^*_{ab}$ statistics~\cite{wyszecki1982color} are shown in the columns 3-7 of Table 1. We can see that the two algorithms also give very close color errors (which suggests that the difference caused by the filter transmittance difference at the spectrum end is negligible in terms of color errors). Compared to the baseline results, we can conclude that by using such optimized filters, we can effectively reduce the color errors by two thirds to three quarters.

\subsubsection{Convergence:} 
\begin{figure}[ht]
	\centering
	\subfloat[Vora-Values]{\includegraphics[width=0.7\textwidth]{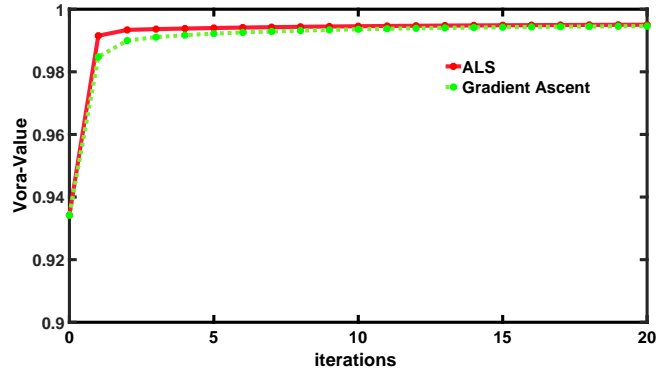}\label{filter1}} \\ \vspace{0.2cm}
	\subfloat[mean $\Delta E_{ab}^*$ ]{\includegraphics[width=0.7\textwidth]{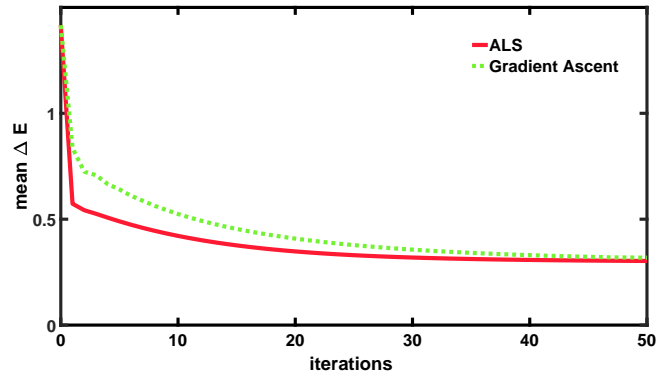}\label{filter2}}	\\
	\caption{Algorithm convergence in terms of Vora-Values and mean $\Delta E_{ab}^*$ with respect to iterations. Solid red line represents the ALS algorithm and the dotted green line represents the gradient ascent algorithm.}
	\label{fig:convergence}
\end{figure}

An important  practical issue in assessing algorithm performance is the convergence speed. In the experiment, we evaluate the filter refinement in each iteration in terms of Vora-Value and averaged mean color error.
In Figure~\ref{fig:convergence}, we show how the two algorithms converge in terms of iterations. In red, we show the results solved by ALS algorithm while the dotted green for that solved by gradient ascent. We see here that both algorithms converges quickly just about 20 iterations for Vora-Value (about 50 iterations for color error) and converge to the same optimal target. Comparatively, ALS algorithm converges much quicker than the gradient ascent algorithm in terms of both metrics.

\section{Conclusion}\label{sec_conclusion}
Previous work has shown that by the addition of a specially designed transmittance filter, a camera can become significantly more colorimetric either by better satisfying the Luther condition~\cite{CIC2018} or by optimizing   the Vora-Value score~\cite{CIC2020}. For a given camera, however, these two methods find different filters that have different performances.

In  this  paper,  we unify the filter design method by making a simple modification to the prior art of the Luther-condition filter design. We propose to find the filter that together with a linear transform best maps the camera sensitivities to an   orthonormalized  set of the color matching functions.  Most importantly, we prove that by using the orthonormal basis, the Luther-condition based optimization becomes equivalent to the Vora-Value filter optimization.
The optimal filter is solved by using the simple alternating least-squares (ALS) algorithm. Significantly, the ALS approach converges more quickly compared to the gradient ascent algorithm previously used for the Vora-filter. Experiments validate the proposed method solved by a simpler algorithm delivering almost the same results compared to the prior art of the Vora-Value optimization.

\bibliographystyle{splncs04}
\bibliography{mybib}

\begin{thebibliography}{10}
\providecommand{\url}[1]{\texttt{#1}}
\providecommand{\urlprefix}{URL }
\providecommand{\doi}[1]{https://doi.org/#1}

\bibitem{barnard2002data}
Barnard, K., Martin, L., Funt, B., Coath, A.: A data set for color research.
  Color Research \& Application  \textbf{27}(3),  147--151 (2002)

\bibitem{berns2019billmeyer}
Berns, R.S.: Billmeyer and Saltzman's principles of color technology. John
  Wiley \& Sons, 4 edn. (2019)

\bibitem{2019CCIW}
Finlayson, G.D., Zhu, Y.: Finding a colour filter to make a camera colorimetric
  by optimisation. In: International Workshop on Computational Color Imaging.
  pp. 53--62. Springer (2019)

\bibitem{CIC2018}
Finlayson, G.D., Zhu, Y., Gong, H.: Using a simple colour pre-filter to make
  cameras more colorimetric. In: Color and Imaging Conference. pp. 182--186.
  Society for Imaging Science and Technology (2018)

\bibitem{HORN1984}
Horn, B.K.P.: Exact reproduction of colored images. Computer Vision, Graphics,
  and Image Processing  \textbf{26}(2),  135 -- 167 (1984)

\bibitem{itu2002parameter}
ITU, R.: Parameter values for the hdtv standards for production and
  international programme exchange. Recommendation ITU-R BT. 709-5  (2002)

\bibitem{jiang2013space}
Jiang, J., Liu, D., Gu, J., S{\"u}sstrunk, S.: What is the space of spectral
  sensitivity functions for digital color cameras? In: 2013 IEEE Workshop on
  Applications of Computer Vision (WACV). pp. 168--179. IEEE (2013)

\bibitem{luther1927gebiet}
Luther, R.: Aus dem gebiet der farbreizmetrik. Zeitschrift Technische Physik
  \textbf{8},  540--558 (1927)

\bibitem{neugebauer1956quality}
Neugebauer, H.: Quality factor for filters whose spectral transmittances are
  different from color mixture curves, and its application to color
  photography. JOSA  \textbf{46}(10),  821--824 (1956)

\bibitem{schanda2007colorimetry}
Schanda, J.: Colorimetry: understanding the CIE system. John Wiley \& Sons
  (2007)

\bibitem{vazquez2014perceptual}
Vazquez-Corral, J., Connah, D., Bertalm{\'\i}o, M.: Perceptual color
  characterization of cameras. Sensors  \textbf{14}(12),  23205--23229 (2014)

\bibitem{vora1993measure}
Vora, P.L., Trussell, H.J.: Measure of goodness of a set of color-scanning
  filters. Journal of the Optical Society of America A  \textbf{10}(7),
  1499--1508 (1993)

\bibitem{vora1997design}
Vora, P.L., Trussell, H.J.: Mathematical methods for the design of color
  scanning filters. IEEE Transactions on Image Processing  \textbf{6}(2),
  312--320 (1997)

\bibitem{wyszecki1982color}
Wyszecki, G., Stiles, W.S.: Color science: Concepts and Methods, Quantitative
  Data and Formulae. Wiley New York, 2 edn. (1982)

\bibitem{zhang2001rank}
Zhang, T., Golub, G.H.: Rank-one approximation to high order tensors. SIAM
  Journal on Matrix Analysis and Applications  \textbf{23}(2),  534--550 (2001)

\bibitem{2020EI}
Zhu, Y., Finlayson, G.: An improved optimization method for finding a color
  filter to make a camera more colorimetric. In: Electronic Imaging 2020.
  Society for Imaging Science and Technology (2020)

\bibitem{CIC2020}
Zhu, Y., Finlayson, G.D.: Designing a color filter via optimization of
  vora-value for making a camera more colorimetric. arXiv:2005.06421 [cs.CV]
  (2020)

\end{thebibliography}

\end{document}